\documentclass[11pt,a4paper]{article}
\usepackage{graphicx}
\usepackage{amsfonts}
\usepackage{booktabs}
\usepackage{siunitx}
\usepackage{multirow}
\usepackage{rotating}
\usepackage{hhline}
\usepackage{dsfont}
\usepackage{amsmath}
\usepackage{amssymb}
\usepackage{graphicx}
\usepackage{latexsym}
\usepackage{marvosym}
\usepackage{comment}
\usepackage{tabularx}             
\usepackage{mathtools}
\usepackage{lipsum}     
\usepackage{authblk}
\usepackage{algorithmic}
\usepackage[ruled,vlined,linesnumbered]{algorithm2e}

\newcommand{\ones}{\mathcal{P}}
\newcommand{\zeros}{\mathcal{N}}
\newcommand{\unbound}{\mathcal{U}}

\newcommand{\freq}{\textsc{FreqRare}\xspace}
\newcommand{\dist}{\textsc{DistanceMIS}\xspace}

\newcommand{\SDB}{\mathcal{D}}
\newcommand{\trans}{\mathcal{T}}			
\newcommand{\items}{\mathcal{I}}
\newcommand{\mis}{\mathcal{S}}
\newcommand{\timeout}{{\sc to}}
\newcommand{\oom}{{\sc oom}}
\def\cp{${\tt CP4MIS}$\xspace}
\def\cfp{${\tt CFPGrowth}$\xspace}
\def\cfpg{${\tt CFPGrowth++}$\xspace}
\def\cfpgp{${\tt CFPGrowth++(PP)}$\xspace}
\def\cfpgc{${\tt CFPGrowth++(C)}$\xspace}
\def\cfpgpr{${\tt CFPGrowth++(CPP)}$\xspace}
\def\msapriori{${\tt MSApriori}$\xspace}
\def\mmsc{${\tt MMS\_Cumulate}$\xspace}
\def\mmss{${\tt MMS\_Stratify}$\xspace}
\def\fpg{${\tt FPGrowth}$\xspace}
\def\fpme{${\tt FP-ME}$\xspace}

\newtheorem{defi}{Definition}
\newtheorem{proposition}{Proposition}

\newtheorem{example}{Example}[section]

\newcommand{\proof}{\noindent \emph{Proof. }}
\newcommand{\qed}{\hfill $\square$}

\usepackage{etoolbox}
\newbool{seeAll}
\usepackage[usenames,dvipsnames]{color}
\usepackage[normalem]{ulem}

\newcommand{\zoo}{{$\tt Zoo$}\xspace}
\newcommand{\vote}{{$\tt Vote$}\xspace}
\newcommand{\anneal}{{$\tt Anneal$}\xspace}
\newcommand{\chess}{{$\tt Chess$}\xspace}
\newcommand{\mushroom}{{$\tt Mushroom$}\xspace}
\newcommand{\connect}{{$\tt Connect$}\xspace}

\newcommand{\tfour}{{$\tt T40$}\xspace}
\newcommand{\pumsb}{{$\tt Pumsb$}\xspace}

\begin{document}

%\title{Mining Frequent Itemsets with Multiple Minimum Supports: a Constraint-Based Approach}

\title{Frequent Itemset Mining with Multiple Minimum Supports: a Constraint-based Approach}

\author[1]{Mohamed-Bachir Belaid}
\author[2]{Nadjib Lazaar}
\affil[1]{Simula Research Laboratory, 
Oslo, Norway}
\affil[2]{LIRMM, University of Montpellier, CNRS, Montpellier, France}
\affil[ ]{\textit {\{bachir@simula.no, lazaar@lirmm.fr\}}}
\date{}

%\author{\IEEEauthorblockN{Mohamed-Bachir Belaid}
%\IEEEauthorblockA{\textit{Simula Research Laboratory} \\
%Oslo, Norway \\
%bachir@simula.no}
%\and
%\IEEEauthorblockN{Nadjib Lazaar}
%\IEEEauthorblockA{\textit{LIRMM, University of Montpellier, CNRS} \\
%Montpellier, France \\
%lazaar@lirmm.fr}
%}

\maketitle

%\vspace*{-.8cm}

%%%%%%%%%%%%%%%%%%%%%%%%%%%%%%%%%%%%%%%%%%%%%%%%%%%%%%%%%%%%%%%%%%%%%%%%%%%%%%%%%%%%%%
%%%%%%%%%%%%%%%%%%%%%%%%%%%%%%%%%%%%%%%%%%%%%%%%%%%%%%%%%%%%%%%%%%%%%%%%%%%%%%%%%%%%%%										ABSTRACT
%%%%%%%%%%%%%%%%%%%%%%%%%%%%%%%%%%%%%%%%%%%%%%%%%%%%%%%%%%%%%%%%%%%%%%%%%%%%%%%%%%%%%%
%%%%%%%%%%%%%%%%%%%%%%%%%%%%%%%%%%%%%%%%%%%%%%%%%%%%%%%%%%%%%%%%%%%%%%%%%%%%%%%%%%%%%%

\begin{abstract}
The problem of discovering frequent itemsets including rare ones 
has received a great  deal of attention.
The mining process needs to be flexible enough to extract frequent and rare regularities at once.
On the other hand, it has recently been shown that constraint programming
is a flexible way to tackle data mining tasks. 
In this paper,  we propose a constraint programming approach for mining itemsets with multiple minimum supports.
Our approach provides the user with the possibility to express any kind of constraints on the minimum item supports.
An experimental analysis
shows the practical effectiveness of our approach compared to the state of the art.

\end{abstract}

%%%%%%%%%%%%%%%%%%%%%%%%%%%%%%%%%%%%%%%%%%%%%%%%%%%%%%%%%%%%%%%%%%%%%%%%%%%%%%%%%%%%%%
%%%%%%%%%%%%%%%%%%%%%%%%%%%%%%%%%%%%%%%%%%%%%%%%%%%%%%%%%%%%%%%%%%%%%%%%%%%%%%%%%%%%%%										Introduction
%%%%%%%%%%%%%%%%%%%%%%%%%%%%%%%%%%%%%%%%%%%%%%%%%%%%%%%%%%%%%%%%%%%%%%%%%%%%%%%%%%%%%%
%%%%%%%%%%%%%%%%%%%%%%%%%%%%%%%%%%%%%%%%%%%%%%%%%%%%%%%%%%%%%%%%%%%%%%%%%%%%%%%%%%%%%%

\section{Introduction}

%\section{Introduction}

Discovering relevant patterns for a particular user remains a challenging task in data mining.
In real-life applications, relevant patterns may be either frequent or rare ones in the data.
%
%
%Mining association rules is one of the most studied problems in data mining.  
%Association Rules (ARs)
%were originally introduced by Agrawal et al. \cite{agrawal1993mining} 
%for sales transactions and products.
%An association rule captures an information of the kind 
%{\em "if we have A and B, the chances to have C are high"} where the frequency of the pattern $ABC$  needs to respect a given frequency (i.e., support) threshold.  
%
In itemset mining, setting the minimum support threshold is a real dilemma (a high value misses 
rare itemsets, a low value generates a large number of meaningless itemsets).
To tackle the \emph{rare item problem} \cite{liu1999mining}, several approaches were proposed 
to mine frequent pattern with multiple minimum supports.
In \cite{liu1999mining}, the problem of mining frequent itemsets with multiple Minimum Item 
Supports (MIS) was introduced with a first revision of Apriori algorithm (\msapriori).
Then, other Apriori-like approaches were proposed like \mmsc{} and \mmss{} \cite{tseng2001mining}.
The well-known \fpg{} was extended with a condensed FP-tree structure to mine frequent itemsets with multiple MIS (\cfp{} \cite{kiran2009improved}, \cfpg{} \cite{kiran2011novel}).
In \cite{gan2017mining}, \fpme{} was proposed based on set-enumeration-tree structure and \emph{sorted downward closure} property.   
%
%\cite{gan2017mining,hahsler2006model,han1995discovery,hu2006mining,kiran2009improved,kiran2010mining,kiran2010towards,kiran2011novel,liu1999mining,ryang2014discovering,selvi2009mining,wang2000mining,yun2003mining,zhou2007association}.
%In \cite{liu1999mining} every item is associated with a minimum support according to its nature. The algorithm \emph{MSApriori} is then proposed to handle this framework. 
%Then, in \cite{kiran2011novel} an algorithm based on \emph{FP-Growth} was proposed.  This algorithm is known to be one of the most efficient for this setting. Some method followed-up like FP-ME \cite{gan2017mining} or MIS-Eclat \cite{darrab2017vertical}.
%In \cite{han1995discovery} every level (itemset with a given size) is associated with a minimum support or/and a minimum confidence. 
%In \cite{wang2000mining} support constraints are defined according to bins of items.
%For thorough overviews, refer to \cite{weiss2004mining,fournier2017survey}.\mbb{To Do}

The specialized algorithms introduced previously are effective for mining patterns with multiple MIS.
However, most of the time the user is interested in patterns that satisfy some specific properties. 
For instance, the user may ask for patterns where items are around the same frequency threshold, or patterns of an adaptive size (large frequent patterns and/or concise rare patterns).
Looking for patterns with additional user-specified constraints remains a bottleneck. 
According to Wojciechowski and Zakrzewicz \cite{wojciechowski2002dataset}, 
there are three ways to handle the additional user's constraints.
We can use a pre-processing step that restricts the dataset to only transactions
that satisfy the constraints. 
Such a technique cannot be used on all kinds of constraints.
We can use a post-processing step to filter out the patterns violating the user's constraints. 
Such a brute-force technique can be computationally
infeasible when the problem without the user's constraints has too many solutions. 
We can finally
integrate the filtering of the user's constraints into the specialized data mining
process in order to extract only the patterns satisfying the constraints. 
Such a technique requires the development of a new algorithm for each new mining problem with user's constraints.

In a recent line of work \cite{de2008constraint,lazaar2016global,schaus2017coversize,belaid2019constraint,belaid2019constraint2},
constraint programming (CP) has been used as 
a declarative way to solve some data mining tasks.
%, such as itemset mining 
%or sequence mining. 
Such an approach has not competed yet with state of the art 
data mining algorithms in terms of CPU time for standard data mining queries 
but the CP approach is competitive as soon as we need 
to add user's constraints. In addition, adding constraints is easily 
done by specifying the constraints directly in the model
without the need to revise the solving process. 
However, the CP approach has not yet been applied to mining frequent itemsets with multiple MIS.

In this paper we introduce a CP model for finding frequent itemsets with multiple MIS.
For that, we introduce a new global constraint, \freq for mining frequent itemsets with multiple MIS.  
We provide a propagator for \freq{} and we show that, for a given variable ordering,
it is sufficient to mine frequent itemsets with multiple MIS in backtrack-free manner.
Our constraint, \freq{}, can be used to express any kind of constraints on the minimum item supports.
We show that our CP model can easily be extended for taking into account any kind of user's constraints.
Experiments on several known large-scale datasets show the effectiveness of our CP model.

The paper is organized as follows.
Section~\ref{sec:background} gives some background material. 
Section~\ref{sec:cpmis} presents our global constraint \freq and its propagator.
Section~\ref{sec:constrained} presents the possible extensions of our CP model to express constraints on the minimum item supports.
Section~\ref{sec:exps} reports experiments. 
Finally, we conclude in Section~\ref{sec:conclusion}.
%%%%%%%%%%%%%%%%%%%%%%%%%%%%%%%%%%%%%%%%%%%%%%%%%%%%%%%%%%%%%%%%%%%%%%%%%%%%%%%%%%%%%%
%%%%%%%%%%%%%%%%%%%%%%%%%%%%%%%%%%%%%%%%%%%%%%%%%%%%%%%%%%%%%%%%%%%%%%%%%%%%%%%%%%%%%%										Background
%%%%%%%%%%%%%%%%%%%%%%%%%%%%%%%%%%%%%%%%%%%%%%%%%%%%%%%%%%%%%%%%%%%%%%%%%%%%%%%%%%%%%%
%%%%%%%%%%%%%%%%%%%%%%%%%%%%%%%%%%%%%%%%%%%%%%%%%%%%%%%%%%%%%%%%%%%%%%%%%%%%%%%%%%%%%%
\section{Background}
\label{sec:background}
\subsection{Itemset mining}
\label{subsec:itemset}

Let $\items = \{p_1,\ldots,p_n\}$ be a set of $n$ distinct objects, called \emph{items}. 
An \emph{itemset} $P$ is a non-empty subset of $\items$. 
%The set of all possible  itemsets is denoted by $\mathcal{L}_{\items}$, that is, 
%$\mathcal{L}_{\items} = 2^{\items}\setminus \varnothing$. 
A \emph{transactional dataset} $\SDB$ is a bag of $m$ itemsets 
$t_1,\ldots,t_m$, called \emph{transactions}. 
The \emph{cover} of an itemset $P$ in $\SDB$, denoted by $cover(P)$, 
is the bag of transactions from $\SDB$ containing $P$.

The \emph{frequency} of an itemset $P$ in $\SDB$, denoted by $freq(P)$, 
is the cardinality of its cover, i.e. $freq(P)=|cover(P)|$. 
Let $\mis=\{s_1,\ldots,s_n\}$ be a set of minimum supports associated to items (i.e., the multiple MIS set), where $s_i$ is the minimum support of the item $p_i$. The itemset $P$ is frequent iff: $$freq(P) \geq \min\limits_{p_i\in P}s_i$$

For the sake of simplicity, we replace in what follows items $p_i$ and transactions $t_j$ with their respective indices $i$ and $j$ and 
we denote the presence of item $i$ in transaction $j$ by $\SDB_{ij}$.

%%%%%%%%%%%%%%%%%%%%%%%%%%%%%%%%%%%%%%%%%%%%%%%%%%%%%%%%%%%%%%%%%%%%%%%%%%%%%%%%%%%%%
%%%%%%%%%%%%%%%%%%%%%%%%%%%%%%%%%%%%%%%%%%%%%%%%%%%%%%%%%%%%%%%%%%%%%%%%%%%%%%%%%%%%%%										Table1: Example
%%%%%%%%%%%%%%%%%%%%%%%%%%%%%%%%%%%%%%%%%%%%%%%%%%%%%%%%%%%%%%%%%%%%%%%%%%%%%%%%%%%%%%
%%%%%%%%%%%%%%%%%%%%%%%%%%%%%%%%%%%%%%%%%%%%%%%%%%%%%%%%%%%%%%%%%%%%%%%%%%%%%%%%%%%%%%

\begin{example}\label{exa: dataset}

The $D$ dataset  presented in  Table \ref{Tab:example} has $4$ items and $5$ transactions.
According to its multiple MIS, $AB$ is frequent, $ABC$ is infrequent and $ABCD$ is frequent.
% $AC$ is infrequent 
%For the itemset $AB$ The cover of $AB$ is $cover(AB) = \{1, 3, 5\}$. Its frequency is the cardinality of its cover, i.e. $freq(AB) = |cover(AB)| = 3$.
%The itemset $AC$ is infrequent, $freq(AC) = 2$ and $\min\limits_{i\in AC}MIS(i) = 3 > freq(AC)$. On the other hand,  the itemset $ACD$ is frequent, $freq(ACD) = 1$ and $\min\limits_{i\in ACD}MIS(i) = 1 \leq freq(ACD)$.
\begin{table}[h]
	\caption{\small Transaction dataset example $D$ with its corresponding MIS set $S$.}\label{Tab:example}
		\centering
		
		$D$:  \begin{tabular}{|c|cccc|}
			\hline
			trans. & \multicolumn{4}{c|}{Items}        \\ \hline
			$t_1$ & $A$ & $B$ & $ $ & $D$ \\ 
			$t_2$ & $A$ & $ $ & $C$ & $D$ \\ 
			$t_3$ & $A$ & $B$ & $C$ & $D$ \\ 
			$t_4$ & $ $ & $B$ & $C$ & $ $ \\
			$t_5$ & $A$ & $B$ & $C$ & $ $\\ \hline
			\end{tabular}
			\quad\quad\quad\quad $S:$ 
		   \begin{tabular}{|c|c|c|c|}
			\hline
			$A$ & $B$ & $C$ & $D$\\ \hline
			4 & 3 & 3 & 1\\  \hline
				
		\end{tabular}
\end{table}

\end{example}

\subsection{Constraint programming}
\label{subsec:cp}

A \emph{Constraint Programming  model} (or \emph{CP  model}) 
specifies a set of variables $X=\{x_1,\ldots, x_n\}$, 
a set of domains $dom=\{dom(x_1),\ldots, dom(x_n)\}$, 
where $dom(x_i)$ is the finite set 
of possible values for $x_i$, 
and a set of constraints $\mathcal{C}$ on $X$.
A constraint $c_j \in \mathcal{C}$ is a relation that specifies the allowed 
combinations of values for its variables $var(c_j)$. 
An assignment 
%(aka, instantiation) 
on a set $Y\subseteq X$ of variables 
is  a mapping from variables in $Y$ to values, and a valid 
assignment is an assignment where all  values belong to the domain of their variable. 
A solution is an assignment on $X$
satisfying all constraints. 
%The constraint satisfaction problem (CSP) consists in deciding whether 
%an instance of a CP model has solutions (or in finding a solution). 
Constraint programming  is the art of writing 
problems as CP models and solving them by finding solutions.  
Constraint solvers typically use backtracking search to explore the search
space of partial assignments. At each assignment, constraint propagation 
algorithms
(aka, propagators) 
prune the search space by enforcing local consistency
properties such as domain consistency.
% (complete propagation).

%A constraint $c_j$ on $X(c_j)$ is \emph{domain consistent (DC)} \mbb{Backtrack-free?} if and only if, for every  $x_i \in X(c_j)$ and every
%$v \in dom(x_i)$, there is a valid assignment satisfying $c_j$ such
%that $x_i = v$. 

\emph{Global constraints} 
are constraints defined by a relation on a non-fixed number of variables.
These constraints allow the solver to better capture the structure of the 
problem. 
The constraint {\sc AllDifferent}, specifying that all its variables must take different values is an example of global constraint
%Examples of global constraints are {\sc AllDifferent}, {\sc Regular},  {\sc Among}, etc.
%For instance, the constraint $\textsc{AllDifferent}(x_1,\ldots,x_n)$ ensures that the variables from
%  $x_1$ to $x_n$ should take different values where $n$ can be any number in $\mathbb{N^*}\setminus \{1\}$  
  (see \cite{rossi2006handbook}).

\begin{example} \label{example-csp} 
Consider  the following instance of a CP model. 
$X= \{x_1, $ $x_2, x_3\}$, 
$dom(x_1) = \{0,2\}$,
$dom(x_2) =\{0,2,4\}$, 
$dom(x_3) =\{1,2,3,4\}$, and 
$\mathcal{C} = \{x_1 \geq x_2,x_1 + x_2 = x_3\}$. 
Value 4 for $x_2$ will be removed by  domain consistency because of constraint  $x_1 \geq x_2$. 
Values 1 and 3 for $x_3$ will be removed by  domain consistency because of constraint  $x_1 + x_2 = x_3$. 
This  instance of CP model  admits the two solutions   $(x_1 = 2, x_2 = 0, x_3 = 2)$
and $(x_1 = 2, x_2 = 2, x_3 = 4)$. 
\end{example}

%\subsection{Declarative itemset mining}
%\label{subsec:declatativeim}
%Most declarative methods use Boolean variables for representing itemsets, 
%where $x_i$ represents the presence of the item $i\in \items$ in the itemset. 
%$x$ is a vector of Boolean variables, $(x_1,\ldots, x_{|\items|})$ to represent the solution. 
%
%Given a vector of Boolean variables $x$, we use the following notations:
%\begin{itemize}
%	\item $\ones=\{i \in \items \mid  dom(x_i)=\{1\}\}$. 
%	\item $\zeros=\{i \in \items \mid  dom(x_i)=\{0\}\}$.
%	\item $\unbound=\{i \in \items \mid dom(x_i)=\{0, 1\}\}$.\\
%\end{itemize} 
%
%Note that $\ones\cup \zeros \cup \unbound = \items$ and $\ones\cap \zeros \cap \unbound = \varnothing$.\\
%
%\begin{example}
%Given a dataset with $4$ items, i.e. $\items = \{A, B, C, D\}$, the instantiation $x = [1, 0, 1, 0/1 ]$ corresponds to the sets $\ones = \{A,C\}$, $\zeros = \{B\}$ and $\unbound = \{D\}$.
%\end{example}

\subsection{CP model for itemset mining} \label{sec:cp4im}

In \cite{de2008constraint}, De Readt et al. have introduced CP4IM, a first CP model to solve itemset mining tasks.
For mining frequent itemsets, the CP model uses two vectors of Boolean variables $x$ and $y$.
$x_i$ represents the presence of item $p_i$ in the searched itemset.
$y_j$ represents the presence of the searched itemset in the transaction $t_j$.
Bear in mind that $x$ are decision variables representing the searched itemset, where $y$ are auxiliary variables 
representing the cover of the searched itemset.
For mining frequent itemsets and given a unique minimum support $s$, the CP model is expressed using two sets of reified constraints:

\begin{equation}\label{eq:channeling}
	\forall j \in \SDB: (y_j=1) \leftrightarrow \sum_{i\in \items} x_i(1-D_{ij})=0
\end{equation}  

\begin{equation}\label{eq:freq}
\forall i \in \items: (x_j=1) \rightarrow \sum_{j\in \SDB} y_j D_{ij}\geq s
\end{equation}  
where,
\begin{itemize}
	\item[(1)] are channelling constraints of arity $(n+1)$ ensuring the relationship between  $x$ and $y$.
	\item[(2)] are constraints of arity $(m+1)$ ensuring the minimum frequency of the searched itemset wrt a minimum support $s$.
	  
\end{itemize}

\newcommand{\reified}{${\tt ReifiedMIS}$}
\section{CP for frequent itemsets with multiple MIS}
\label{sec:cpmis}

Thanks to the expressiveness of CP, the CP4IM model presented in Section \ref{sec:cp4im} can easily be revised to mine 
frequent itemsets with multiple MIS. 
All we have to do is to replace the constraints (\ref{eq:freq}) by (\ref{eq:mis}):

\begin{equation} \label{eq:mis}
 \forall i\in \items: (x_i =1) \rightarrow  \sum\limits_{j\in \SDB} y_j D_{ij}\geq \min\limits_{k\in \items}(s_k: x_k\neq 0)
\end{equation}

where the minimum operator in (\ref{eq:mis}) returns the corresponding minimum item support $s_i$
during search. 
Let us call the revised version of CP4IM for multiple MIS the CP model \reified{}.
The \reified{} model represents a straightforward CP encoding of the problem using reified constraints.\footnote{A reified constraint connects a constraint to a boolean variable that catches its truth value \cite{rossi2006handbook}.} 
Such encoding suffers from scalability issue due to the use of auxiliary variables $y$ and an important number of reified constraints.
\reified{} requires $(n+m)$ constraints of arity $(n+1)$ and $(m+1)$ to encode the whole dataset.

For an effective and scalable CP model, we propose the \freq{} global constraint encoding the minimum frequency constraint with multiple MIS (equation (\ref{eq:mis})).
\freq{} requires neither reified constraints nor auxiliary variables.

Our global constraint is expressed only on the decision variables $x$.
We will use the following notations:
\begin{itemize}
	\item $\ones=\{i \in \items \mid  dom(x_i)=\{1\}\}$. 
	\item $\zeros=\{i \in \items \mid  dom(x_i)=\{0\}\}$.
	\item $\unbound=\{i \in \items \mid dom(x_i)=\{0, 1\}\}$.\\
\end{itemize}

\begin{defi}[\freq{} Constraint]
Let $x$ be a vector of Boolean variables.
Let $\SDB$ be a dataset and $\mis$ the corresponding set of minimum item supports (MIS).
The global constraint $\freq_{\SDB, \mis}(x)$ holds if and only if: $$(\unbound=\varnothing) \wedge (freq(\ones)\geq \min\limits_{p_i\in \ones}s_i)$$ 
\end{defi}

\begin{example}
	Consider the transaction dataset of Table \ref{Tab:example}.
	Let $x=\langle x_A, x_B$ $,x_C, x_D\rangle$ with $dom(x_i)=\{0,1\}$ for $i\in\{A,B,C,D\}$.
	Consider the frequent itemset $CD$ encoded by $x=\langle 0,0,1,1\rangle$,
	where $\ones=\{C,D\}$, $\zeros=\{A,B\}$ and $\unbound=\varnothing$.
	Here, $\freq_{D, S}(x)$ holds because all variables are instantiated (i.e., $\unbound=\varnothing$) 
	and $[freq(CD)=2]\geq [\min(s_C, s_D)=1]$.\\
\end{example}
%
%\mbb{To chose between both definitions}
%
%\begin{defi}[\freq{} Constraint]
%Let $x$ be a vector of Boolean variables,  $MIS$ a list of minimum item supports and $\SDB$ a dataset.	
%An assignment on $x$ and $s$ satisfies $\freq_{\SDB, MIS}(x)$ if and only if $freq(\ones)\geq min_{i\in \ones}[MIS(i)]$. 
%\end{defi}

%\subsection{Filtering algorithm}

%In the following we present a filtering algorithm for $\freq_{\SDB, MIS}(x)$

%%%%%%%%%%%%%%%%%%%%%%%%%%%%%%%%%%%%%%%%%%%%%%%%%%%%%%%%%														   Algorithm 
%%%%%%%%%%%%%%%%%%%%%%%%%%%%%%%%%%%%%%%%%%%%%%%%%%%%%%%%%

We present now a propagator algorithm for the global constraint \freq.\\
   
\subsubsection*{Algorithm.} 
Algorithm~\ref{alg:filter} takes as input the variables $x$ and the multiple item supports $\mis$.  	
We start by computing the 
cover of the itemset $\ones$ and store it in ${\tt cover}$ (line~\ref{lineAlg:cover}). 
Then, we compute the minimum item support $s\in \mis$ wrt $\ones$ and $\unbound$ items (line~\ref{lineAlg:lb}).
If $\ones$ is infrequent wrt $s$, then $\ones$ cannot be extended to a solution  (using $\unbound$ items) and we return a failure (line~\ref{lineAlg:fail}). 
Otherwise, we must remove items $p_i\in \unbound$ that cannot belong to a solution containing $\ones$
%If yes, no super-set of $\ones$ can be frequent and a fail is returned (line~\ref{lineAlg:fail}). 
%Algorithm \ref{alg:filter} then filters value $1$ from $x_i$ if including the item $i$ would result a frequency that is less than $s$ 
(lines~\ref{lineAlg:loop1}-\ref{lineAlg:unbound}).

\begin{algorithm}
	
	\caption{ Propagator for $\freq$}\label{alg:filter} 
	\SetKw{Input}{In}
	\SetKw{InOut}{InOut}	
	\SetKw{In}{In}
	\SetKw{Out}{Out}
	\SetKw{InOut}{InOut}	
	\SetKw{Output}{Output}
	\SetKw{return}{return}
	
	\DontPrintSemicolon
	\BlankLine
	%\LinesNotNumbered
	\In: $\mis$: minimum item supports\;
	\InOut:
	$x = \{x_1,\ldots, x_n\}$: Boolean item variables;
	\BlankLine
	
	\Begin{
		${\tt cover} \gets cover(\ones);$\\ \label{lineAlg:cover}
		$s \gets \min\limits_{i\in \ones \cup \unbound}s_i;$	\label{lineAlg:lb}
		
	    \If{$|{\tt cover}| < s$\label{lineAlg:failc}}	{
		\Return \emph{failure}; \label{lineAlg:fail}
		}		
		%\tcp{Filtering on item variables}
		\ForEach{$i\in \unbound$\label{lineAlg:loop1}	}	{				
			\If{$|{\tt cover} \cap cover(i)| < s$ \label{lineAlg:check} }	
				{$dom(x_i) \gets dom(x_i)\setminus\{1\}$\;\label{lineAlg:filter}	
					 $\unbound \gets \unbound\setminus\{i\};$ \label{lineAlg:unbound}	
					 
				 	 $\zeros \gets \zeros\cup\{i\};$} \label{lineAlgzeros}		}						
		}	
		
\end{algorithm}

%\subsection{Discussion}

\newcommand{\minmis}{${\tt minMis}$}

\begin{proposition}[Backtrack-free search]\label{prop:free}
	Enumerating frequent itemsets with multiple MIS using the propagator of \freq (Algorithm~\ref{alg:filter}) with an increasing minimum item support as variable ordering heuristic: \minmis{} heuristic, is backtrack-free.
	
%	
%	Using the strategy of search that starts with $x_i$ such that $MIS(i)$ is the smallest, the propagator in Algorithm~\ref{alg:filter} is backtrack-free on the constraint \freq. 
\end{proposition}

\proof
%
%At line \label{lineAlg:lb}, Algorithm~\ref{alg:filter} computes the minimum support corresponding to the current instantiation where it corresponds to the minimum MIS value of $\ones \cup \unbound$ items.
%Here, $s$ wi 
%The search process to enumerate the frequent itemsets with multiple MIS is done with an increasing minimum item support as variable selector (\minmis{} heuristic) and using \freq{} at each node of the search space.
%Suppose that $x_i$ is the first variable to select (i.e., $s_i$ is minimal in $\mathcal{S}$).
%If $s_i< s$, then a \emph{failure} is returned at line \ref{lineAlg:fail} proving that no solution exists.
%Knowing that $x_i$ is a variable representing the presence of $p_i$ in the returned itemset, as $p_i$ is infrequent,
%any superset of $p_i$ is infrequent too using \minmis{} heuristic (anti-monotony of the frequency).
%Suppose now that for a given partial instantiation $\ones$ we have a cover greater than $s_i\geq s$ (no failure with $\ones=\{p_i\}$),
%and a superset of $p_i$ is infrequent.
%Here,  then Algorithm~\ref{alg:filter} will never return a \emph{failure} state as $p_i$ is frequent and any superset of $p_i$ 
%is a frequent itemset using \minmis{} heuristic (anti-monotony of the frequency).

%  
%we call the propagator of \freq{} with a partial instantiation where $\ones\neq \varnothing$ 

We first prove that if \freq admits a solution,  $\ones$ is necessarily one of them. 
Suppose there is a solution and $\ones$ is not one of them. 
This means that there exits a superset of $\ones$,  $\ones\cup i$, which is frequent.
As we use \minmis{} heuristic, we have the guarantee that $s$, the minimum support computed at line~\ref{lineAlg:lb}, 
corresponds to an item in $\ones\cup \unbound$ (i.e., not in $\zeros$). 
That is,  $$s = \min\limits_{j\in \ones\cup \unbound}(s_j) = \min\limits_{j\in ((\ones\cup\{i\})\cup (\unbound\setminus\{i\}))}(s_j)$$

Here, if $\ones\cup i$ is frequent wrt $s$, $\ones$ is necessary frequent (anti-monotony of the frequency) and thus $\ones$ is a solution too,  which contradicts the
assumption.

We now prove that Algorithm~\ref{alg:filter} returns failure
if and only if \freq does not admit any solution.
We know that \freq has no solution if and only if whatever the partial instantiation submitted to Algorithm~\ref{alg:filter},
 $\ones$ is not a solution.
 No solutions means that whatever the item $i$ its frequency is below the corresponding minimum item support $s_i$  (i.e., $\forall i \in \items: freq(i)< s_i$).
 Here, the search process will prune the value $1$ from the domain of all $x_i$ at line~\ref{lineAlg:filter} and thus, converge on a complete instantiation where $\zeros=\mathcal{I}, \ones=\unbound=\varnothing$.
 
% 
 % 
%In such a case the test at line~\ref{lineAlg:failc} is true and a \emph{failure} state is reached.
%If \freq admits solutions,  $\ones$ is a support for
%$x_i = 0$, for all $i\in \unbound$. 
%As a result, Algorithm~\ref{alg:filter} does not
%need to check consistency of value 0 for any variable.
%
We now prove that Algorithm~\ref{alg:filter} prunes value $1$ from
$dom(x_i)$ exactly when $i$ cannot belong to a solution containing $\ones$.
Suppose that value $1$ of $x_i$
is pruned by Algorithm~\ref{alg:filter}.
This means that the test in line~\ref{lineAlg:check} was true, that
is,  $\ones\cup \{i\}$ is infrequent.
Thus, and by definition,  $\ones\cup \{i\}$ does not belong to any solution. 
Suppose now that value $1$ of $x_i$
is not pruned.  
From line~\ref{lineAlg:check},  we deduce that $\ones\cup \{i\}$ is frequent.
Thus $\ones\cup \{i\}$ is a solution
and mining frequent itemsets with multiple MIS is backtrack-free using Algorithm~\ref{alg:filter} with \minmis{} heuristic.
\qed\\

\begin{proposition}[Time complexity of Algorithm~\ref{alg:filter}] 
	Given a transaction dataset $\SDB$ of $n$ items and $m$ transactions and $\mis$ the corresponding set of minimum item supports (MIS),
	Algorithm~\ref{alg:filter} has an 
	$O(n\times m)$ time complexity.
\end{proposition}
\proof
%\mbb{To Do.}
Computing the size of the cover of an itemset is in 
$O(n\times m)$ (line~\ref{lineAlg:cover}). 
The cover of $\ones\cup\{i\}$ in line~\ref{lineAlg:check} is computed in $O(m)$ thanks to ${\tt cover}$.
The loop in line~\ref{lineAlg:loop1} is called at most $n$ times leading to a time complexity of lines~\ref{lineAlg:loop1}-\ref{lineAlg:filter} in $O(n\times m)$.
Thus, Algorithm~\ref{alg:filter} has an $O(n\times m)$  time complexity.
\qed\\

\section{Constrained  frequent itemsets with multiple MIS}
\label{sec:constrained}

In this section, we illustrate the power of CP to state diverse queries, while maintaining the declarativeness of our CP model and by taking 
into account different user-specified constraints.\\     

In addition to mining frequent itemsets with multiple MIS, the user may have more restrictions on the itemsets to mine.
In this section we discuss the possible constraints that a user may have on the set of MIS.\\
Let $Q_0$ refers to the basic query: \emph{"mining frequent itemsets with multiple MIS"}. 

\subsection{Distance between MISs} \label{sec:distance}
The user may be interested in itemsets that include items of the same nature. 
That is, distances between MISs are bounded above by a given value $ub$.
%any two MIS in the itemset should be less than an upper bound.
For instance, in a sales transactions dataset, the item \emph{\{car\}} occurs rarely and should have a low MIS value (e.g.,  $s_{car}  = 1$).
On the other hand,  the item \emph{\{bread\}} occurs frequently and should have a high MIS value (e.g.,  $s_{bread}  = 1,000$).
In this case, the itemset \emph{\{car, bread\}} is frequent wrt $s_{car}$.
To avoid generating such itemset, we can put a restriction on the distance between MIS pairs, for instance,  $(|s_{i} - s_{j}|\leq 50: \forall i,j$),
where in such case, the itemset \emph{\{car, bread\}} will not be a relevant one to return.

In our CP model, the distance constraint can be expressed as follows:

$$\dist_{\mis,ub}(x)\equiv \forall i,j\in\items, \  (|s_i-s_j| \ x_i \ x_j\ \leq ub)$$

The user may ask the following query:

\definecolor{Gray}{gray}{0.85}
\newcolumntype{Y}{>{\centering\arraybackslash}X}
\newcolumntype{Z}{>{\raggedleft\arraybackslash}X}
\newcolumntype{a}{>{\columncolor{Gray}}c}
\begin{center}
	\begin{tabularx}{\linewidth}{|Y|}
		\hline
		\textit{}\\
		\textit{$Q_1:$ Given  a dataset $\SDB$, an MIS vector $\mis$ and an upper bound $ub$,  }
		%	\textit{ }\\
		\textit{extract frequent itemsets of MIS distances bounded above by $ub$.}\\
		\textit{}\\
		\hline
	\end{tabularx}
\end{center}

The query $Q_1$ can easily be  expressed in CP using \freq  and with the user distance constraints as follows:
\[
Q_1 (\SDB, \mis, ub)=\begin{cases}
	\freq_{\SDB, \mis}(x) & \\
	\dist_{\mis,ub}(x) &\\
\end{cases}
\]

%The role of each constraint is the following:
%\begin{itemize}
%	\itemsep0ex
%	\item[(4)] ensures that the itemset $\ones$ is frequent wrt $\mis$.
%	\item[(5)] ensures that the distance between MIS pairs is not exceeding an upper bound $ub$.
%\end{itemize}

\subsection{Cardinality constraint} \label{sec:cardinality}
In addition to the distance constraint, the  user may ask to strengthen her query with a restriction on the cardinality of the returned itemsets.
For instance,  the user may ask for itemsets with a size of at least $c$.

Here a query that the user may ask:

\begin{center}
	\begin{tabularx}{\linewidth}{|Y|}
		\hline
		\textit{}\\
		\textit{$Q_2:$ Given  a dataset $\SDB$, an MIS vector $\mis$,  }
			\textit{ an upper bound $ub$  and a lower bound $c$, }
		\textit{extract frequent itemsets of MIS distances bounded above by $ub$}
		\textit{and of a size of at least $c$.}\\
		\textit{}\\
		\hline
	\end{tabularx}
\end{center}

The query $Q_2$ can be expressed in CP as follows:

\[
Q_2 (\SDB, \mis, ub,c)=\begin{cases}
Q_1(\SDB, \mis, ub) &\\
\sum\limits_{i\in \items} x_i\geq c &\\

\end{cases}
\]
%
%\begin{itemize}
%	\itemsep0ex
%	\item[(1)] ensures that the itemset $\ones$ is frequent wrt $\mis$ and  distances between MIS pairs are not exceeding an upper bound $ub$ (see Section \ref{sec:distance}).
%	\item[(2)] ensures that the itemset is of a size of at least $c$.
%\end{itemize}

\subsection{$k$-patterns mining}
\label{subsec:kpttern}
A promising road to discover useful patterns is to impose constraints on a set of $k$ related patterns ($k$-pattern sets) \cite{guns2011k}. 
In this setting, the interest of a pattern is evaluated wrt a set of patterns.
%The user may also be interested in exactly $k$ itemsets with relation between them \cite{guns2011k}.
For instance, the user may be interested in exctracting $k$ distinct frequent itemsets with a cardinality and distance constraints on MIS.

%distance between MIS pairs note exceeding an upper bound $ub$ and a size of at least $c$. 

Here an example of a particular query that the user may ask:

\begin{center}
	\begin{tabularx}{\linewidth}{|Y|}
		\hline
		\textit{}\\
		\textit{$Q_3:$ Given  a dataset $\SDB$,  an integer $k$, an MIS vector $\mis$, }\
			\textit{ an upper bound $ub$  and a lower bound $c$,}
		\textit{extract $k$ distinct frequent itemsets of MIS distances bounded above by $ub$}
		\textit{and with a sizes of at least $c$.}\\
		\textit{}\\
		\hline
	\end{tabularx}
\end{center}

The query $Q_3$ can be expressed as follows:

\[
Q_3 (\SDB,\mis,k, ub,c)=\begin{cases*}
\forall i\in[1,k]: \freq_{\SDB, \mis}(x^{i}) & (1)\\
\forall i \in[1,k]: \dist_{\mis,ub}(x^{i})&(2)\\
\forall i \in[1,k]: \sum\limits_{j\in \items} x^{i}_j\geq c&(3)\\
\forall i,j\in[1,k]: x^{i}\cap x^{j} = \varnothing&(4)\\
\end{cases*}
\]

The role of each type of constraint is the following:
\begin{itemize}
\itemsep0ex
\item[(1)] ensures that for every $i$ in $[1,k]$, the itemset $\ones^i$ is frequent wrt $\mis$. 
\item[(2)] ensures that distances between MIS pairs of the $k$ itemsets  are not exceeding an upper bound $ub$.
\item[(3)] ensures that the $k$ itemsets are of a size of at least $c$.
\item[(4)] ensures that the $k$ itemsets are distinct. 
\end{itemize}

\section{Experiments}
\label{sec:exps}

This section describes the experimental settings (including benchmark datasets, protocol and implementation), 
the experimental results and comparison with the state of the art approach \cfpg.

\newcommand{\minMIS}{${\tt MIS_{min}}$}
\newcommand{\maxMIS}{${\tt MIS_{max}}$}
\newcommand{\avgMIS}{${\tt MIS_{avg}}$}

\subsection{Experimental protocol}
We selected several real-sized datasets from the FIMI 
repository.\footnote{fimi.ua.ac.be/data/} 
These datasets have various characteristics representing 
different application domains.
The first part of Table~\ref{tab:characteristics} reports, for each dataset, 
the number of transactions $|\mathcal{T}|$, 
the number of items $|\items|$, the average size of transactions 
$\overline{|\mathcal{T}|}$ and the 
density $\rho$ (i.e., $\overline{|\mathcal{T}|}/|\items|$).
The datasets are presented by increasing  size $|\items|\cdot|\mathcal{T}|$.
We  selected datasets of various size and density. 
Some datasets, such as  \zoo{}  and \chess{},  are very dense  
(resp. $44\%$ and $49\%$). 
Others are very sparse (e.g., $4\%$ for \tfour{}). 
The sizes of these datasets vary  from around $4,000$ to more than $10^8$.  

For our experiments, we assign the MIS values for items according to their frequencies  and using the formula proposed in \cite{liu1999mining}:

\begin{equation*}
	s_i = max(\beta \ freq(i),\ {\tt MIS_{min}})
\end{equation*}

where $\beta \in [0,1]$ is a parameter, $freq(i)$ is the frequency of the item $i$ and \minMIS{} is the lowest support that an item can have.
The second part of Table \ref{tab:characteristics} reports, for each dataset, the selected $\beta$ value, the lowest, the highest and the averaged MIS relative values (\minMIS, \maxMIS, \avgMIS).
%For each  dataset, an  instance is characterized by its lowest support $ls$ and the selected $\beta$.  
%For instance, \zooI{5}{0.5} denotes the instance of the \zoo{} dataset 
%with a lowest support of $5$ and $\beta$ of $0.5$.

The implementation of our CP model with $\freq$ global constraint, coined \cp, is carried out in the ${\tt Oscar}$ solver using Scala.\footnote{bitbucket.org/oscarlib/oscar/}
The code is publicly available at ${\tt github.com/CP4MIS/cp4mis}$.
After a few preliminary tests and based on the findings presented in Section \ref{sec:cpmis}, we decided to use \minmis{} as variable ordering heuristic and \emph{largest value first} as value ordering heuristic.
We compared our CP approach to the \cfpg specialized algorithm for extracting frequent itemsets with MIS \cite{kiran2011novel}. 
We used the implementation of \cfpg publicly available in the \textsc{SPMF} platform.\footnote{www.philippe-fournier-viger.com/spmf/}
All experiments were conducted on an Intel core $i7$, 
$2.8Ghz$ with a RAM of $16GB$ and with a timeout of one hour.\\

\newcommand{\RQ}[1]{\textbf{ RQ#1}}

Our evaluation aims to answer the following four research questions:
\begin{itemize}
	\item\RQ{1}: \emph{How effective is the use of \freq global constraint comparing to the basic \reified{} CP model?}
	\item\RQ{2}: \emph{How effective is \cp{} for mining frequent itemsets with multiple MIS (queries of type $Q_0$) and compared to specialized algorithm?}
	\item\RQ{3}: \emph{How effective is \cp{} for mining constrained frequent itemsets with multiple MIS (queries of type $Q_1$ and  $Q_2$) and compared to specialized algorithm?}
	\item\RQ{4}: \emph{How effective is our CP approach for mining $k$ distinct constrained frequent itemsets with multiple MIS (queries of type $Q_3$) and compared to specialized algorithm?}
\end{itemize}
% and  a time limit of one hour.
%%%%%%%%%%%%%%%%%%%%%%%%%%%%%%%%%%%%%%%%%%%%%%%%%%%%%%%%%														 Table2: Dataset Characteristics
%%%%%%%%%%%%%%%%%%%%%%%%%%%%%%%%%%%%%%%%%%%%%%%%%%%%%%%%%

%\begin{table*}[tb]
%	\caption{Dataset Characteristics.} \label{tab:characteristics}
%	\centering
%	\begin{tabularx}{\textwidth}{|l|Z|Z|Z|Y||Z|Z|Z|Z|} \hline
%		%		& &  &  &  & \\
%		{Name} & \multicolumn{1}{c|}{$|\mathcal{T}|$} & \multicolumn{1}{c|}{$|\mathcal{I}|$} & \multicolumn{1}{c|}{$\overline{|\mathcal{T}|}$} & \multicolumn{1}{c||}{$\rho(\%) $}  & $\beta$ & \minMIS & \maxMIS & \avgMIS\\
%		%				& &  &  &  & \\
%		\hline
%		\hline
%		{\zoo}  & 101 & 36 & 16 & 44 &0.1& 1\% & 9\% & 4\%\\
%		{\vote}  & 435 & 48 & 16 & 33 &0.1& 0.2\% & 6\% & 3\%\\
%		{\anneal}  & 812 & 89 & 42 & 45 &0.9& 12\% & 90\% & 45\%\\
%		{\chess}  & 3,196 & 75 & 37 & 49 & 0.5& 31\% & 50\% & 36\%\\
%		{\mushroom}  & 8,124  & 112 & 23  & 19 &0.1& 0.1\% & 10\% & 2\%\\
%		{\connect}  & 67,557  & 129 & 43  & 33  & 0.7 & 30\% & 70\% & 39\%\\
%		%{\tt T10I4D100K}  & 100,000  & 870 & 10  & 1  &0.1& 1 & 783 & 116\\
%		{\tfour}  & 100,000  & 942 & 40  & 4 & 0.1 & 0.05\% & 3\% & 0.5\%\\
%		{\pumsb}  & 49,046  & 2,113 & 74  & 3 &0.9 &41\%& 90\% & 42\%\\
%		%{\retail}  & 88,162 & 16,470 & 10  & 0.06  &0.5& 2 &25338 & 28\\
%		\hline
%		%			\multicolumn{6}{r}{\tone = T10I4D100K $\quad$ \tfour = T40I10D100K  }  \\ 
%		
%	\end{tabularx}
%
%\end{table*}

\begin{table}[tb]
	\caption{Dataset Characteristics.} \label{tab:characteristics}
	\centering
	\begin{tabular}{|l|c|c||c|c|c|c|} \hline
		%		& &  &  &  & \\
		\multirow{ 2}{*}{Name} & \multirow{ 2}{*}{$(|\mathcal{T}|, |\mathcal{I}|)$}  & \multicolumn{1}{c||}{$\rho $}  & \multirow{ 2}{*}{$\beta$} & \minMIS & \maxMIS & \avgMIS\\
		&  & $(\%)$ & & $(\%)$& $(\%)$& $(\%)$ \\
		%				& &  &  &  & \\
		\hline
		\hline
		{\zoo}  & (101,  36)  & 44 &0.1& 1 & 9 & 4\\
		{\vote}  & (435,  48)  & 33 &0.1& 0.2 & 6 & 3\\
		{\anneal}  & (812,  89)  & 45 &0.9& 12 & 90 & 45\\
		{\chess}  & (3K,  75)  & 49 & 0.5& 31 & 50 & 36\\
		{\mushroom}  & (8K,   112)   & 19 &0.1& 0.1 & 10 & 2\\
		{\connect}  & (68K,   129)   & 33  & 0.7 & 30 & 70 & 39\\
		%{\tt T10I4D100K}  & 100,000  & 870 & 10  & 1  &0.1& 1 & 783 & 116\\
		{\tfour}  & (100K,   942)   & 4 & 0.1 & 0.05 & 3 & 0.5\\
		{\pumsb}  & (49K,   2K)   & 3 &0.9 &41& 90 & 42\\
		%{\retail}  & 88,162 & 16,470 & 10  & 0.06  &0.5& 2 &25338 & 28\\
		\hline
		%			\multicolumn{6}{r}{\tone = T10I4D100K $\quad$ \tfour = T40I10D100K  }  \\ 
		
	\end{tabular}
	
\end{table}

\subsection{Results}
\newcommand{\sol}{${\tt \#sol}$}
\newcommand{\node}{{\tt \#Nodes}}
\newcommand{\fail}{{\tt \#Fails}}

Our first experiment compares our \cp model to \reified{} model and to \cfpg{} on queries of type $Q_0$, where the user is looking for frequent itemsets with multiple MIS.
Table \ref{tab:q0} reports the result of the comparison  
with the CPU time given in seconds (s), the memory consumption in megabytes (MB) of the two CP models,  
and the number of solutions \sol{} for each instance.

\subsubsection{\RQ{1}: \reified{} vs \freq{} on $Q_0$}
The main observation when comparing the reified CP model to \cp
is that the use of \freq{} global constraint outperforms significantly the basic model.
Let us take a closer look to the four first datasets where neither a timeout nor an out of memory are reported.
In terms of CPU time, we can observe a speed-up factors of $2$, $9$, $10$ and $98$. 
In terms of memory consumption, we denotes factors of $47$, $63$, $188$ and $270$.
This is explained by the huge number of reified constraint to propagate and to check at each node of the search tree
comparing to a single call per node of \freq propagator.
We denote two timeout  and two out-of-memory instances for \reified{} model.
Again, this is explained by the size of the model and the number of the posted reified constraints.
For instance, if we take \pumsb{} dataset, \reified{} contains $(|\items|+|\trans|)= 51K$ variables,
($|\trans|=49K$) reified constraints to express the channelling constraints on ($|\items|=2K$) variables (equation (\ref{eq:channeling})),
($|\items|=2K$) reified constraints to express the minimum frequency of itemsets on ($|\trans|=49K$) variables (equation (\ref{eq:freq})).
This means that the CP solver has to load in memory a CP model of $51K$ reified constraints expressed on $51K$ variables.
We observe that with a timeout of one hour and a memory of $16GB$, ${\tt Oscar}$ solver is not able to manage CP models of size exceeding 
$(|\items|\times|\trans|)\approx 10^5$.
Another observation is the experimental validation of Proposition \ref{prop:free}.
\freq{} of \cp combined with \minmis{} heuristic enumerates the whole set of frequent itemsets of the  $8$ datasets in a backtrack-free manner and without 
any fail during search (${\tt\#failures}=0$).

\subsubsection{\RQ{2}: \cfpg{} vs \cp{} on $Q_0$}

As expected, \cfpg is very efficient in enumerating all frequent itemsets with multiple MIS.
 \cfpg is from $2$ to $18$ times faster than \cp.
However, our \cp approach is also reasonable and can enumerate up to millions frequent itemsets in a few seconds.
Furthermore, \cp is even better than \cfpg on one instance  i.e. \tfour.

\begin{figure}[h]
\centering 
%\begin{tabular}{cc}
	\includegraphics[width=\linewidth]{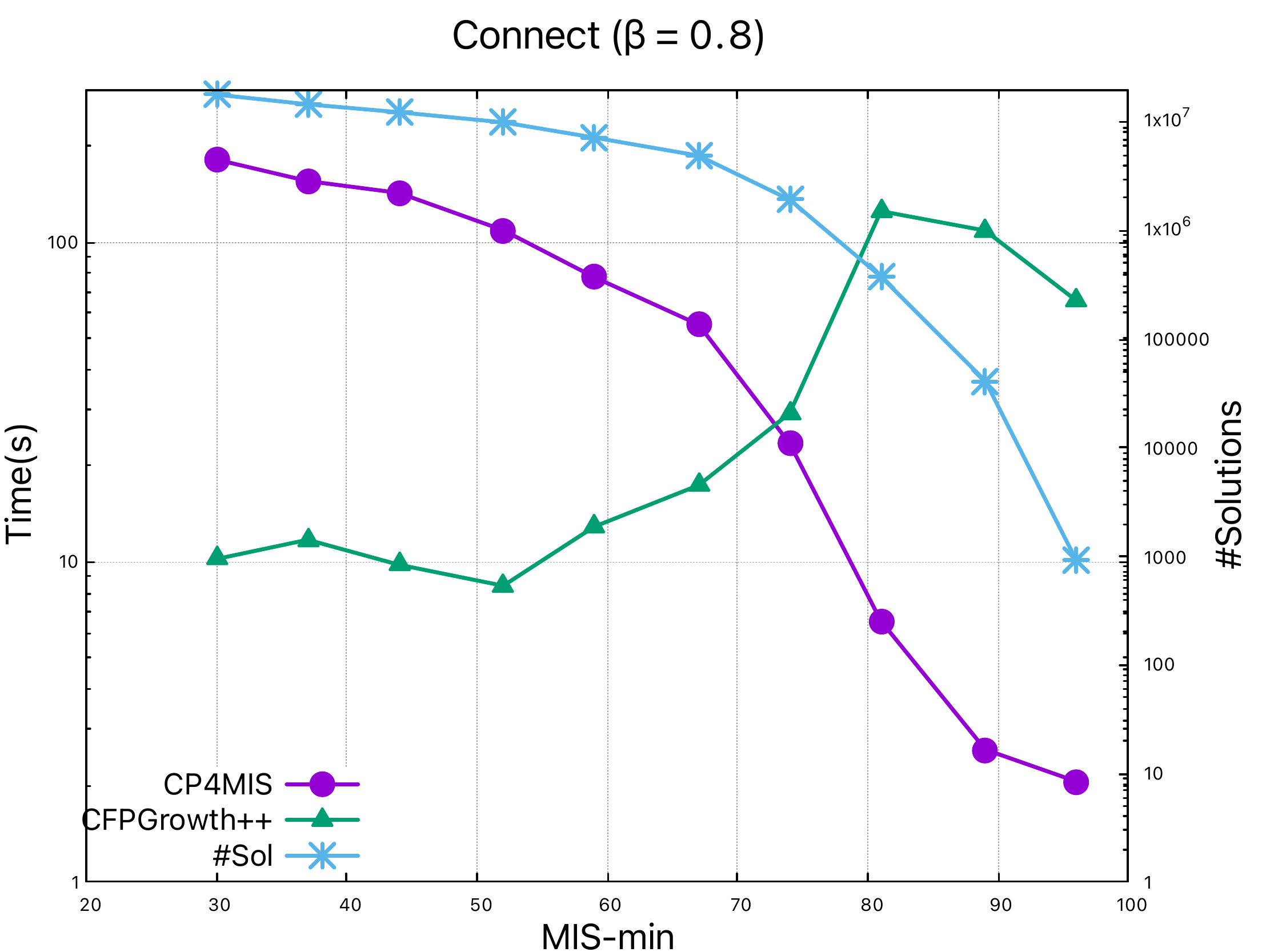} %& 	\includegraphics[scale = 0.25]{Plot-Pumsb-09-Solutions.pdf}\\
%		\textbf{(a)} & \textbf{(b)}
%\end{tabular}
% \caption{(a) \cfpg vs \cp varying  \minMIS on \connect; (b)  \cfpg vs \cp varying \minMIS on \pumsb.}\label{fig:Vary}
	\caption{\cfpg vs \cp varying  \minMIS\ on \connect with $\beta = 0.8$.}\label{fig:varyMin}
\end{figure}

Interestingly, the less solutions there is to mine the better is our approach \cp.
This is illustrated in Figure \ref{fig:varyMin} where we compare \cfpg  to \cp for mining frequent itemsets on \connect with $\beta = 0.8$ and by varying \minMIS.
We clearly see that our approach \cp is highly correlated with the number of solutions compared to \cfpg.
With an \minMIS \ of $30\%$, \cfpg is able to mine more than $17$ millions of itemsets in a few seconds (exactly $10.21$~seconds), whereas \cp spends more than $3$ minutes to generate the whole set of solutions.
However, with \minMIS{} starting from $74\%$, \cp starts to be more efficient and it is able to extract the $915$ itemsets  (${\tt MIS_{min}} = 96\%$) in $2$ seconds, an instance on which \cfpg spends more than one minute.
The explanation for this good behavior of \cp is the strength of constraint propagation 
to rule out inconsistent parts of the search space. On
the contrary, on an instance with $17$ millions solutions ( \minMIS{} = $30\%$ ), 
the CP solver is almost reduced to an enumerating process.

%%%%%%%%%%%%%%%%%%%%%%%%%%%%%%%%%%%%%%%%%%%%%%%%%%%%%%%%%%%%%%%%%%%%%%%%%%%%%%%%%%%%%
%%%%%%%%%%%%%%%%%%%%%%%%%%%%%%%%%%%%%%%%%%%%%%%%%%%%%%%%%%%%%%%%%%%%%%%%%%%%%%%%%%%%%%										Table3: Results on Q0
%%%%%%%%%%%%%%%%%%%%%%%%%%%%%%%%%%%%%%%%%%%%%%%%%%%%%%%%%%%%%%%%%%%%%%%%%%%%%%%%%%%%%%
%%%%%%%%%%%%%%%%%%%%%%%%%%%%%%%%%%%%%%%%%%%%%%%%%%%%%%%%%%%%%%%%%%%%%%%%%%%%%%%%%%%%%%

\begin{table*}
	\caption{\cfpg vs  (\reified{} vs \cp) on the query $Q_0$ (Time in seconds and Memory in MB)} \label{tab:q0}
	\centering
	\scalebox{1}{
		\begin{tabularx}{\linewidth}{|c||Y||Y|c||Y|c||c|} \hline
			
			\multirow{ 2}{*}{\textbf{Dataset}} &(a)  & \multicolumn{2}{c||}{(b)}  &  \multicolumn{2}{c||}{(c)} &  \multirow{2}{*}{\sol}\\ 
			 \cline{2-6}
			 & Time &   Time & Memory  & Time & Memory  & \\
			\hline
			
			%\zoo  & \textbf{0.73} & 1.64 & 1,076,519 \\
			\zoo  & \textbf{0.81} & 12.00 & 3,760 & 1.34 & \textbf{20} & 1,314,983 \\
			%\vote   & \textbf{0.98} & 2.13 & 1,094,559  \\
			\vote   & \textbf{1.56} & 196.17& 2,164& 2.23  &\textbf{8} & 2,177,409  \\
			\anneal & \textbf{30.91} & 134.74& 3,095 & 64.82 & \textbf{49}  &  71,757,451 \\
			 \chess  & \textbf{11.64} & 305.03 &3,153 & 28.20 & \textbf{67} &  22,660,643 \\
			\mushroom  & \textbf{45.53} & \timeout & --  & 106.00  & \textbf{48} & 105,291,573 \\
			\connect  & \textbf{48.45} &  \timeout & --  & 854.59 & \textbf{218} & 91,740,453 \\
			%\tone  &\textbf{4.59} &  246.22 &  1,923,259 \\
			%\tone & \textbf{3.33} & 6.72 & 804,843  \\	
			\tfour  & 409.55 & -- & \oom & \textbf{91.70} & \textbf{2,304} & 15,859,400 \\
			%\pumsb & 186.90 & \textbf{90.76} & 1,897,479 \\
			\pumsb & \textbf{38.60} & -- & \oom & 115.67 & \textbf{916} &  13,507,227 \\
			%\retail  & \textbf{3.11} & 361.74 &  1,241,269\\	
			\hline
			 \multicolumn{7}{r}{(a):\cfpg{}, (b):\reified{}, (c):\cp{}.}\\
			 			%\pumsbI{10000}{0.5}  &  &  &  & \retailI{2}{0.1}  & \textbf{304.66} & \timeout &  508,484,778\\	    
					
		\end{tabularx}
	}
\end{table*}

%%%%%%%%%%%%%%%%%%%%%%%%%%%%%%%%%%%%%%%%%%%%%%%%%%%%%%%%%%%%%%%%%%%%%%%%%%%%%%%%%%%%%
%%%%%%%%%%%%%%%%%%%%%%%%%%%%%%%%%%%%%%%%%%%%%%%%%%%%%%%%%%%%%%%%%%%%%%%%%%%%%%%%%%%%%%										Figure1: Varing Min-MIS
%%%%%%%%%%%%%%%%%%%%%%%%%%%%%%%%%%%%%%%%%%%%%%%%%%%%%%%%%%%%%%%%%%%%%%%%%%%%%%%%%%%%%%
%%%%%%%%%%%%%%%%%%%%%%%%%%%%%%%%%%%%%%%%%%%%%%%%%%%%%%%%%%%%%%%%%%%%%%%%%%%%%%%%%%%%%%

\subsubsection{\RQ{3}: \cfpg{} vs \cp{} on $Q_1$ and  $Q_2$}
Our second experiment compares \cfpgc to our approach, \cp, for mining constrained frequent itemsets (queries of type $Q_1$ and $Q_2$).\\
\cfpgc (${\tt C}$ is for checker) is a revised version of \cfpg with a checker to filter out itemsets violating the constraints.

In Table \ref{tab:q1}, we report the time, in seconds, for the two approaches and for each instance for mining frequent itemsets of MIS distances bounded above by $ub$ (queries of type $Q_1$).
We also report the number of solutions for each instance: \sol.
We selected an $ub$ value for each instance in order to have less than $10K$ solutions.

The main observation that we can draw from Table \ref{tab:q1} is that \cp outperforms \cfpgc{} on all the instances.
Comparing to the results of $Q_0$, the fact that we strengthen the query with the distance constraint makes the CP resolution more effective.
For instance, on \connect, \cp needed around $14$ minutes to enumerate the $91$ millions of solutions corresponding to $Q_0$, where it took only $2$ seconds to return the $1,683$ solutions corresponding to $Q_1$ query.
%drastically improves its performance from more than $15$ minutes for mining the $91$ million frequents itemsets, to only $2$ seconds for extracting an interesting subset (the $1,683$ satisfying $Q_1$). 
Thanks to constraint propagation that drastically reduces the search space.
On the other hand, \cfpgc has no pruning power during the enumeration, and does not take advantage of the additional constraints (the distance constraint in this case) to reduce the search space ($21$ seconds to enumerate the $1,683$ solutions).

To strengthen our findings, we conduct an experiment with $Q_2$ query type. 
In addition to the distance constraint, $Q_2$ considers the cardinality constraint i.e.,   itemsets should have a size of at least $c$.
Table \ref{tab:q2} reports the time, in seconds, for the two approaches and for each instance acting on $Q_2$ query type.
We also report the number of solutions of each instance: \sol.
The parameter $c$ is selected in order to have less than $50$ solutions per instance.

Again, \cp is the winner where it is from 4 to $43$ faster than\\
 \cfpgc.
% wins on every single instance. 
%This time, \cp is even faster to mine the few number of itemsets. 
This is explained by the additional power added to the propagation process during the resolution to enumerate the few solutions present in the huge search space.
%In fact the additional cardinality constraint provides the constraint solver with more pruning and thus reduces the search space.
Where \cfpgc needs to enumerate the millions of candidates and then filter out the ones violating the user-constraints.
%non relevant   has no pruning power and generates all the itemsets to filter most of them at the end.

%%%%%%%%%%%%%%%%%%%%%%%%%%%%%%%%%%%%%%%%%%%%%%%%%%%%%%%%%%%%%%%%%%%%%%%%%%%%%%%%%%%%%
%%%%%%%%%%%%%%%%%%%%%%%%%%%%%%%%%%%%%%%%%%%%%%%%%%%%%%%%%%%%%%%%%%%%%%%%%%%%%%%%%%%%%%										Table4: Results on Q1
%%%%%%%%%%%%%%%%%%%%%%%%%%%%%%%%%%%%%%%%%%%%%%%%%%%%%%%%%%%%%%%%%%%%%%%%%%%%%%%%%%%%%%
%%%%%%%%%%%%%%%%%%%%%%%%%%%%%%%%%%%%%%%%%%%%%%%%%%%%%%%%%%%%%%%%%%%%%%%%%%%%%%%%%%%%%%

\begin{table*}
	\caption{\cfpgc vs  \cp on the query $Q_1$ (time in seconds)} \label{tab:q1}
	\centering
	\scalebox{1}{
	\begin{tabularx}{\linewidth}{|X|Y||Y||Y||Y|} \hline
			
			\textbf{Dataset} & $ub$ &  (a) & (b) & {\tt \#sol}\\ \hline \hline
			
			%\zoo  & 1 & & \textbf{0.12} & 668 \\
			\zoo  & 2 & 0.61& \textbf{0.17} & 5,765 \\
			%\vote   & 1 &  & \textbf{0.20} & 2,466  \\
			\vote   & 1 & 0.82 & \textbf{0.18} & 2,466  \\
			%\anneal & 10&  & \textbf{0.23} &  134\\
			\anneal & 30& 12.95 & \textbf{0.25} & 1,790\\
		     %\chess  & 10 &  & \textbf{0.29} &  91\\
		     \chess  & 80 &5.96 & \textbf{0.27} &  1,442\\
			%\mushroom  & 10 & & \textbf{0.60} &  420\\
			\mushroom  & 50 & 20.20& \textbf{0.46} &  2,641\\
			\connect  & 1000 & 21.28&  \textbf{2.11} & 1,683  \\
%			\connect  & 1000 & &  \textbf{2.08} & 1,683  \\
			%\tone & 1 &  \textbf{3.39}&  93.45 &  915\\	
			\tfour  & 100 & 401.93 &  \textbf{107.92} & 5,846\\
			%\pumsb & 100 &  &\textbf{12.80}  & 127 \\
			\pumsb & 1000 & 26.97&\textbf{3.04}  & 1,287 \\
			%\retail  & 1 & \textbf{2.37} &{\sc oom} & 14,340\\	
			
			%\pumsbI{10000}{0.5}  &  &  &  & \retailI{2}{0.1}  & \textbf{304.66} & \timeout &  508,484,778\\	    
			\hline		
			\multicolumn{5}{r}{(a):\cfpgc{}; (b):\cp{}.}
		\end{tabularx}
	}
\end{table*}

%%%%%%%%%%%%%%%%%%%%%%%%%%%%%%%%%%%%%%%%%%%%%%%%%%%%%%%%%%%%%%%%%%%%%%%%%%%%%%%%%%%%%
%%%%%%%%%%%%%%%%%%%%%%%%%%%%%%%%%%%%%%%%%%%%%%%%%%%%%%%%%%%%%%%%%%%%%%%%%%%%%%%%%%%%%%										Table5: Results on Q2
%%%%%%%%%%%%%%%%%%%%%%%%%%%%%%%%%%%%%%%%%%%%%%%%%%%%%%%%%%%%%%%%%%%%%%%%%%%%%%%%%%%%%%
%%%%%%%%%%%%%%%%%%%%%%%%%%%%%%%%%%%%%%%%%%%%%%%%%%%%%%%%%%%%%%%%%%%%%%%%%%%%%%%%%%%%%%

\begin{table*}
	\caption{\cfpgc vs  \cp on the query $Q_2$ (time in seconds)} \label{tab:q2}
	\centering
	\scalebox{1}{
		\begin{tabularx}{\linewidth}{|X|Y|Y||c||Y||Y|} \hline
			
			\textbf{Dataset} & $ub$ & $c$& (a) & (b) & {\tt \#sol}\\ \hline \hline
			
			%\zoo  & 1 & & \textbf{0.12} & 668 \\
			\zoo  & 2 & 10 &0.62 & \textbf{0.10} & 14 \\
			%\vote   & 1 &  & \textbf{0.20} & 2,466  \\
			\vote   & 1 &10 & 1.14 & \textbf{0.13} & 12 \\
			%\anneal & 10&  & \textbf{0.23} &  134\\
			\anneal & 30 & 8 & 12.78  & \textbf{0.18} &  7\\
		     %\chess  & 10 &  & \textbf{0.29} &  91\\
		     \chess  & 80 & 8 & 6.49 & \textbf{0.23} &  30\\
			%\mushroom  & 10 & & \textbf{0.60} &  420\\
			\mushroom  & 50 & 8 &19.19 & \textbf{0.36} &  27\\
			\connect  & 1000 & 10 &20.88 &  \textbf{2.03} & 2  \\
%			\connect  & 1000 & &  \textbf{2.08} & 1,683  \\
			%\tone & 1 & 5 & \textbf{3.21 }&  44.25 &  0\\	
			%\tone & 1 & 2 & \textbf{ 3.29} &  46.26 & 45\\	
			%\tfour  & 1 & 10 & 391.35 &  \textbf{53.62} & 0 \\
			\tfour  & 100 & 6 &389.80 & \textbf{54.31}  & 14\\
			%\pumsb & 100 &  &\textbf{12.80}  & 127 \\
			\pumsb & 1000 & 8 & 27.91  &\textbf{2.86}  & 17 \\
			%\retail  & 1 & &  &{\sc oom} & \\	
			
			%\pumsbI{10000}{0.5}  &  &  &  & \retailI{2}{0.1}  & \textbf{304.66} & \timeout &  508,484,778\\	    
			\hline		
			\multicolumn{6}{r}{(a):\cfpgc{}, (b):\cp{}.}
		\end{tabularx}
	}
\end{table*}

\subsubsection{\RQ{4}: \cfpg{} vs \cp{} on $Q_3$.}
For our last experiment, we compares \cfpg to \cp on a query of type $Q_3$ in $k$-patterns mining context.
% (queries of type $Q_3$).   
Table \ref{tab:selected} reports the different instances selected for our experiment.
The instances are selected  in order to vary the number of solutions from $0$ to $100K$ $k$-patterns solutions.
% reasonable number of solutions due to the huge search space for k-patterns mining. 

Using CP, the problem is expressed using $k$ Boolean vectors (see Section \ref{subsec:kpttern}).

For such problem, a baseline can be the use of specialized algorithm like \cfpg combined with a 
post-processing step. 
One can imagine two scenarios:
\begin{enumerate}
	\item  The use of \cfpg to mine the total number of frequent itemsets with multiple MIS and a generate-and-test search trying  to find distinct itemsets satisfying the user-constraints (distance and cardinality constraints in our case). This baseline is coined \cfpgp ({\tt PP} is for post-processing).
	\item The use of \cfpgc to mine the total number of frequent itemsets with multiples MIS satisfying the user-constraints and a generate-and-test search trying to find distinct itemsets. This second baseline is coined \cfpgpr (${\tt CPP}$ is for checker + post-processing).
\end{enumerate}  

Such baselines can be very expensive. 
In both cases, the post-processing will generate the possible $k$ combinations of itemsets. 
Table \ref{tab:selected} reports the solutions of $Q_0$ and $Q_2$ that represent the number of candidates on which the $k$ combinations will be generated for, respectively, \cfpgp and \cfpgpr baselines.

%%On the other hand, \cfpg does not support $k$-patterns mining and is obliged to explore all the search space\footnote{$C_{k}^{n}$.} as a post processing step.
%We have implemented two versions of  \cfpg:
%\begin{itemize}
%\item[$\bullet$] \cfpgp\footnote{PP for post processing.} which apply a post processing step directly on the frequent itemsets (e.g., $5,553$ for \chess).  
%\item[$\bullet$] \cfpgpr which apply \cfpgc to generate constrained frequent itemsets and then apply a post processing step on those (e.g.,  $932$ for \chess).
%\end{itemize}

Table \ref{tab:q3} reports the CPU time, in seconds,  for each instance  and for the three approaches (\cfpgp, \cfpgpr and \cp) acting on $Q_3$ query type.
%for mining $k$ distinct frequent itemsets that satisfy the distance and cardinality constraints (queries of type $Q_3$).
We also report the number of solutions of each instance: \sol.

The main observation that we can draw from Table \ref{tab:q3} is that \cp is able to cope with such complex query
and to return the $96K$ $k$-patterns solutions in $40$ minutes.
This demonstrates again the power of propagation using a CP resolution.
On the other hand, the \cfpgp baseline timeouts on all instances and \cfpgpr on 5 out of the 8 instances.
This is an expected result knowing that the baselines have to cope with a massive number of combinations.
For instance, on \connect{} dataset and $k=6$ we have more than $10^{24}$ candidates. 

If we take the instance of \chess{} with $k=6$.
Here we have an instance without solution and \cp is able to prove it in less than $26$ seconds, where the two baselines reach timeout without proving the unsatisfiability of the instance.

% is thanks to the pruning power of the constraint solver.
%\cfpgp is not able to solve any instance within the time limit.
%This is due to the huge search space that should be explored during the post processing step (e.g., more than $10^{24}$ candidates on \connect with $k = 6$).
Note that the used checker in \cfpgpr allows the baseline to reduce the possibilities, but not enough to avoid the explosion when $k$ grows.
\cfpgpr first prunes non-solutions using \cfpgc reducing the search space and thus solving the \connect instance when $k=6$ in $3$ minutes.
However,  with $k = 7$,  the search space explodes and \cfpgpr is not able to prove no solution exist within the time limit.

%Let us take a closer look to \mushroom with $k=9$. 
%Here, \cp is able to return the first solution in $40$ seconds, where \cfpgp and \cfpgpr are able to extract the first solution in, respectively, XX and YY minutes.
 
\begin{figure}[h]
  \centering 
  \includegraphics[width=\linewidth]{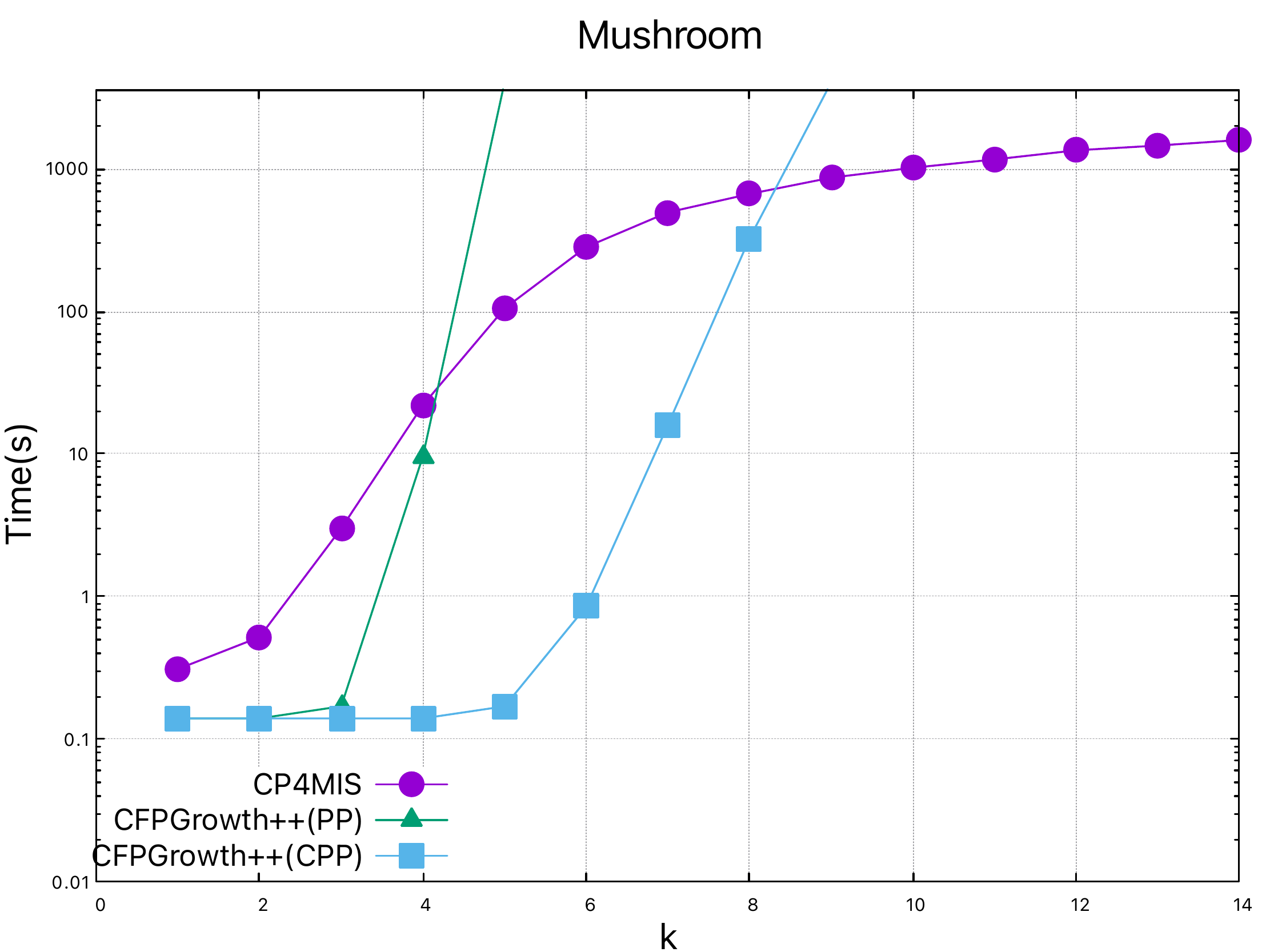}
  \caption{\cfpgp vs \cfpgpr vs \cp for mining k patterns on \mushroom.}\label{fig:kpattern}
\end{figure}

To strengthen our observations,  we compare the three approaches on \mushroom while varying $k$ in Figure \ref{fig:kpattern}.
We observe that the two baselines \cfpgp and \cfpgpr are acting better than \cp within a small $k$ (i.e., $k<5$).
However, \cp scales very well when $k$ grows and it is able to prove that no solution exists on $k = 14$ instance in less than $27$ minutes.
While the two baselines \cfpgp and \cfpgpr follow an exponential scale and they reach the one hour time limit when $k$ is equal to, respectively, 5 and 9. 
% exceed the one hour time limit (starting from $k = 5$ and $k = 9$ respectively).

%%%%%%%%%%%%%%%%%%%%%%%%%%%%%%%%%%%%%%%%%%%%%%%%%%%%%%%%%%%%%%%%%%%%%%%%%%%%%%%%%%%%%
%%%%%%%%%%%%%%%%%%%%%%%%%%%%%%%%%%%%%%%%%%%%%%%%%%%%%%%%%%%%%%%%%%%%%%%%%%%%%%%%%%%%%%										Table6: Selected instances for Kpattern 
%%%%%%%%%%%%%%%%%%%%%%%%%%%%%%%%%%%%%%%%%%%%%%%%%%%%%%%%%%%%%%%%%%%%%%%%%%%%%%%%%%%%%%
%%%%%%%%%%%%%%%%%%%%%%%%%%%%%%%%%%%%%%%%%%%%%%%%%%%%%%%%%%%%%%%%%%%%%%%%%%%%%%%%%%%%%%

\begin{table*}
	\caption{Selected instances for k-patterns mining.} \label{tab:selected}
	\centering
	\begin{tabularx}{\linewidth}{|l|Y|c|c|c||c|c|c|c|} \hline
		%		& &  &  &  & \\
		{Dataset} & $\beta$ & \minMIS & \maxMIS & \avgMIS &${\tt \#sol}_1$ &\multicolumn{1}{c|}{$ub$} &\multicolumn{1}{c|}{$c$} &${\tt \#sol}_2$ \\
		%				& &  &  &  & \\
		\hline
		\hline
		{\chess}   & 0.9& 78\% & 90\% & 80\% & 5,553 & 150 & 3 & 932\\
		{\mushroom}  &0.9& 18\% & 90\% & 26\% & 2,977 & 3000 & 2 & 138\\
		{\connect}  & 0.9 & 89\% & 90\% & 89\% & 41,143 & 300 & 3 & 397\\
		{\pumsb}  &0.9 &88\%& 90\% & 88\% & 7,044& 400 & 3 & 347\\
		
		\hline
		\multicolumn{9}{r}{${\tt \#sol}_1$ = solutions of $Q_0$ $\quad$ ${\tt \#sol}_2$ = solutions of $Q_1$. }  \\ 
		
	\end{tabularx}
	
\end{table*}

%%%%%%%%%%%%%%%%%%%%%%%%%%%%%%%%%%%%%%%%%%%%%%%%%%%%%%%%%%%%%%%%%%%%%%%%%%%%%%%%%%%%%
%%%%%%%%%%%%%%%%%%%%%%%%%%%%%%%%%%%%%%%%%%%%%%%%%%%%%%%%%%%%%%%%%%%%%%%%%%%%%%%%%%%%%%										Table7: Results on Q3
%%%%%%%%%%%%%%%%%%%%%%%%%%%%%%%%%%%%%%%%%%%%%%%%%%%%%%%%%%%%%%%%%%%%%%%%%%%%%%%%%%%%%%
%%%%%%%%%%%%%%%%%%%%%%%%%%%%%%%%%%%%%%%%%%%%%%%%%%%%%%%%%%%%%%%%%%%%%%%%%%%%%%%%%%%%%%

\begin{table*}
	\caption{\cfpgp vs \cfpgpr vs \cp on the query $Q_3$ (time in seconds)} \label{tab:q3}
	\centering
	
		\begin{tabularx}{\linewidth}{|X|Y|Y||Y||Y||Y|} \hline
			
			\textbf{Dataset} & $k$ & (a) & (b) & (c)  &{\tt \#sol}\\ \hline \hline

		     \chess  &5 &  \timeout & 61.41 & \textbf{19.10} & 480\\
		      \chess  &6 &  \timeout &  \timeout & \textbf{25.42} &0\\

			\mushroom  &9 & \timeout  & \timeout & \textbf{862.55} & 1,407\\
			\mushroom  &10 & \timeout  & \timeout & \textbf{1014.49} & 0\\
			\connect  & 6 &\timeout   & 182.71 &  \textbf{133.68} & 36,537\\
			\connect  & 7 &\timeout   & \timeout  &  \textbf{143.69} & 0\\

			\pumsb & 6  & \timeout  & 199.51 & \textbf{177.80} & 61,186  \\
				\pumsb & 7  & \timeout  & \timeout & \textbf{202.49} & 0  \\
	    
			\hline		
			\multicolumn{6}{r}{(a):\cfpgp{}, (b):\cfpgpr, (c):\cp.}
		\end{tabularx}
	
\end{table*}

%%%%%%%%%%%%%%%%%%%%%%%%%%%%%%%%%%%%%%%%%%%%%%%%%%%%%%%%%%%%%%%%%%%%%%%%%%%%%%%%%%%
%%%%%%%%%%%%%%%%%%%%%%%%%%%%%%%%%%%%%%%%%%%%%%%%%%%%%%%%%%%%%%%%%%%%%%%%%%%%%%%%%%%%%%										Figure2: K-pattern
%%%%%%%%%%%%%%%%%%%%%%%%%%%%%%%%%%%%%%%%%%%%%%%%%%%%%%%%%%%%%%%%%%%%%%%%%%%%%%%%%%%%%%
%%%%%%%%%%%%%%%%%%%%%%%%%%%%%%%%%%%%%%%%%%%%%%%%%%%%%%%%%%%%%%%%%%%%%%%%%%%%%%%%%%%%%%

To sum up, our experimental evaluation shows  that a specialized algorithm like \cfp is faster 
on basic queries (e.g., asking for frequent itemsets), but it cannot cope with complex queries in a huge search space and of few solutions. 
It would need to think and to propose ad-hoc solutions,
whereas CP approach enables a novice DM-user to express his query as constraints.

%%%%%%%%%%%%%%%%%%%%%%%%%%%%%%%%%%%%%%%%%%%%%%%%%%%%%%%%%%%%%%%%%%%%%%%%%%%%%%%%%%%%%%
%%%%%%%%%%%%%%%%%%%%%%%%%%%%%%%%%%%%%%%%%%%%%%%%%%%%%%%%%%%%%%%%%%%%%%%%%%%%%%%%%%%%%%										Conclusion
%%%%%%%%%%%%%%%%%%%%%%%%%%%%%%%%%%%%%%%%%%%%%%%%%%%%%%%%%%%%%%%%%%%%%%%%%%%%%%%%%%%%%%
%%%%%%%%%%%%%%%%%%%%%%%%%%%%%%%%%%%%%%%%%%%%%%%%%%%%%%%%%%%%%%%%%%%%%%%%%%%%%%%%%%%%%%
\section{Conclusion}
\label{sec:conclusion}
In this paper, we have introduced a constraint programming approach for itemset mining with multiple minimum supports MIS.
For this, we have defined a new global constraint \freq and provided a filtering algorithm that mine frequent itemsets with multiple MIS in backtrack-free manner,  given a variable ordering.
We have empirically evaluated our CP approach.
The experiments showed the performance of our CP model comparing to the specialized approach,  \cfpg.
Although \cfpg is very efficient on basic queries like mining frequent itemsets, our CP approach outperforms \cfpg on complex queries with user-constraints and/or with search space explosion in a $k$-patterns mining.  
Furthermore, our CP approach provides the user with the flexibility to express any kind of constraints including constraints on MIS without the need to revise the solving process.

\section*{Acknowledgments}
This work has received funding from
the T-LARGO project under grant agreement No 274786.

%\newpage

\bibliography{arxiv}

\bibliographystyle{abbrv}

\end{document}